%% file: root.tex
\documentclass[letterpaper, 10 pt, conference]{ieeeconf}  

\usepackage[utf8]{inputenc} 
\usepackage[T1]{fontenc}    
\usepackage{hyperref}       
\usepackage{url}            
\usepackage{booktabs}       
\usepackage{amsfonts}       
\usepackage{nicefrac}       
\usepackage{microtype}      
\usepackage{xcolor}         

\usepackage{times}
\usepackage{epsfig}
\usepackage{graphicx}
\usepackage{amsmath}
\usepackage{amssymb}
\usepackage{caption,subcaption}
\usepackage{multirow, booktabs}
\usepackage{siunitx}
\usepackage{xspace}
\usepackage{times}
\usepackage{epsfig}

\usepackage{array}
\usepackage{textcomp}
\usepackage{xcolor}
\usepackage{makecell}

\IEEEoverridecommandlockouts                              

\title{\LARGE \bf RenderNet: Visual Relocalization Using Virtual Viewpoints in Large-Scale Indoor Environments
}

\author{Jiahui Zhang, Shitao Tang, Kejie Qiu, Rui Huang, Chuan Fang, Le Cui, Zilong Dong, Siyu Zhu, and Ping Tan
\thanks{All the authors are with Alibaba XRLab. zjh271224@alibaba-inc.com}
}

\begin{document}

\maketitle

\begin{abstract}
Visual relocalization has been a widely discussed problem in 3D vision: given a pre-constructed 3D visual map,
the 6 DoF (Degrees-of-Freedom) pose of a query image is estimated. Relocalization in large-scale indoor environments enables attractive applications such as augmented reality and robot navigation. However, appearance changes fast in such environments when the camera moves, which is challenging for the relocalization system. To address this problem, we propose a virtual view synthesis-based approach, RenderNet, to enrich the database and refine poses regarding this particular scenario. Instead of rendering real images which requires high-quality 3D models, we opt to directly render the needed global and local features of virtual viewpoints and apply them in the subsequent image retrieval and feature matching operations respectively.
The proposed method can largely improve the performance in large-scale indoor environments, e.g., achieving an improvement of 7.1\% and 12.2\% on the Inloc dataset.
\end{abstract}
\input{introduction}
\input{relatedwork}
\input{methods}
\input{experiment}

\input{conclusion}

\bibliography{relocalization}
\bibliographystyle{plain}
\end{document}

%% file: introduction.tex
\section{Introduction}
In order to satisfy the growing demand for various spatial intelligence relevant
applications, such as navigation, Augmented Reality (AR)
and SLAM, visual relocalization is an indispensable technology.
It is essentially a registration problem to localize the query images within a visual database. By establishing 2D-3D correspondences between query images and database images, 6 DoF camera poses of the query images can be estimated.

\begin{figure}[]
 \centering
\includegraphics[width=0.5\textwidth]{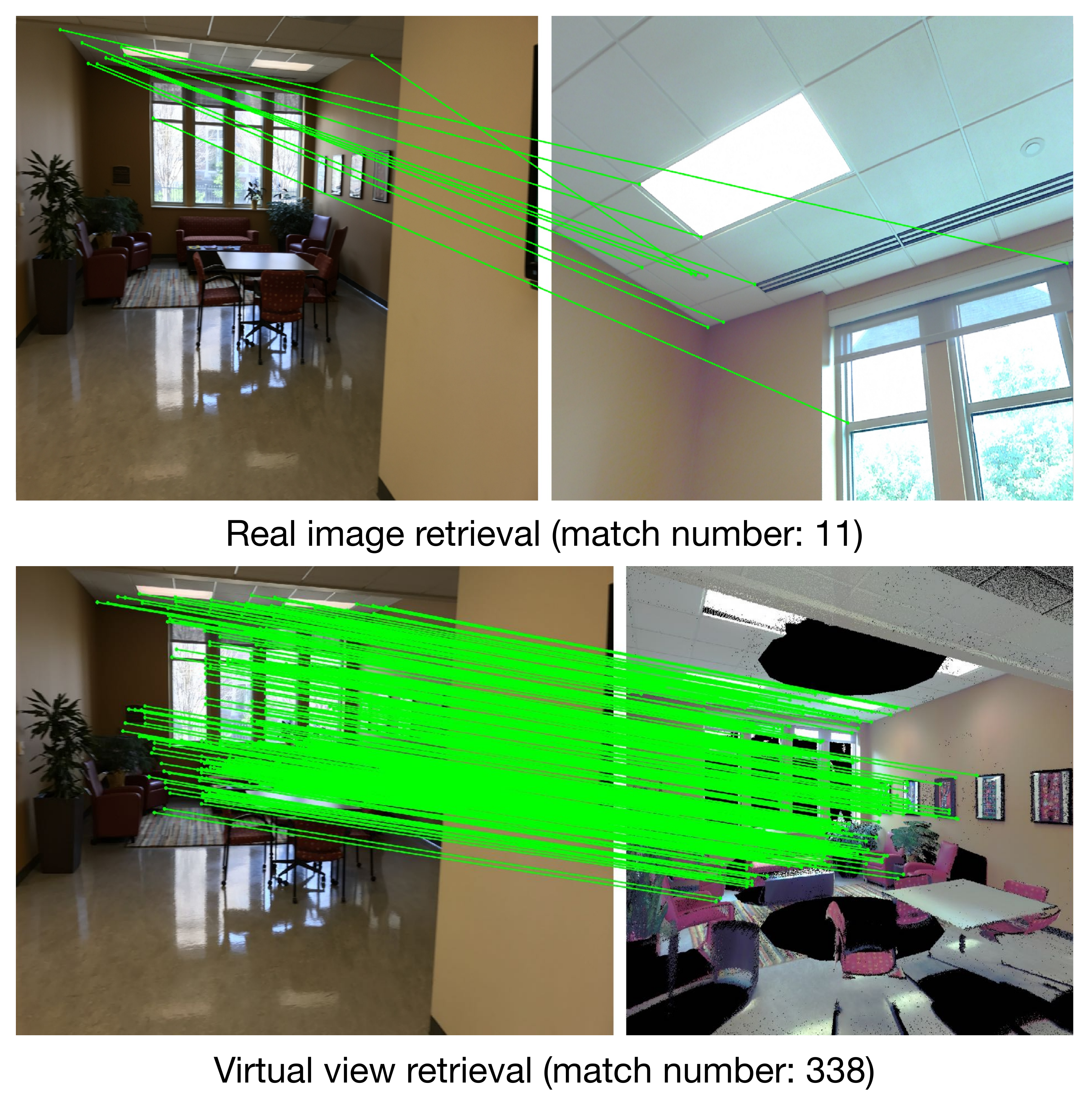}
 \caption{Comparison of the state-of-the-art approach HLoc ~\cite{sarlin2019coarse} and the proposed RenderNet. The top image pair shows retrieved real database image while the bottom image pair shows virtual synthesis views. Putative matches from SuperGlue matcher~\cite{sarlin2020superglue} are shown.
 The virtual view, which is not explicitly generated, can have a larger overlap with the query image than real database images, thus can be helpful for both image retrieval and image matching and result in better relocalization performance.
 }
 \label{fig:motivation}
\end{figure}

Current state-of-the-art scene agnostic localization methods ~\cite{sattler2018benchmarking, sarlin2019coarse} are comprised of an image retrieval module searching for candidate database images as the localization reference, and an image matching module to establish 2D-3D correspondences. Then, the camera poses are computed through a subsequent RANSAC + Perspective-n-Point (PnP) algorithm. 
However, 
the performance of both modules suffers from large view and appearance changes, which are common in large-scale indoor environments, since appearance changes rapidly when the camera moves.
As shown in Fig. \ref{fig:motivation}, the most similar images in the database might still have insufficient overlap with the query image, which are hard to be retrieved and matched, thus resulting in degraded performance.

In order to address this problem, an intuitive idea is to 
synthesis novel views to enrich the database.
Database images with larger overlaps make image retrieval and image matching easier.
Thus the overall relocalization performance can be improved accordingly.
However, image synthesis requires high-quality 3D models, which are usually captured by high-cost 3D scanners and this process is time-consuming. 
For example, the Inloc ~\cite{wijmans2017exploiting} relocalization dataset is captured using an expensive Lidar scanner~\cite{faro} which requires more than 30 minutes for a single scan. 

Instead of rendering novel views from a pre-built high-quality 3D model, we propose RenderNet to generate virtual novel views from the existing database images.
Different from the general view synthesis process~\cite{martin2020nerf, nakata2002synthesis} that aims to render visual-realistic images, the proposed neural network learns to directly obtain global features and local features of virtual viewpoints used in the image retrieval module and image matching module respectively.
We opt to generate the needed features, which are the final quantities for localization, instead of generating images, to 
reduce the influence of inevitable artifacts in the synthesized images.

With the generated virtual views, we optimize the traditional localization pipeline by plugging in two operations, namely the view augmentation operation and pose refinement operation.
In the view augmentation stage, we manually pre-sample a set of possible views in a scene as virtual views, so that query images can retrieve those views with their rendered global features, and correspondences can be established via local features. 
In the pose refinement stage, 
a new virtual view is generated using the coarse camera pose computed in the previous stage.
Then 2D-3D correspondences can be matched and refined poses are obtained subsequently. 

Those generated virtual views can largely improve the image retrieval and correspondences establishing process.
Our method is suitable for indoor scenes where images change rapidly when viewpoints change.
In the experimental section, we will demonstrate that the proposed approach can largely improve the baseline localization method, (\textbf{7.1\% and 12.2\% of accuracy improvement in the two scenes of Inloc}). 

%% file: relatedwork.tex
\section{Related work}
\paragraph{Visual relocalization.} Visual relocalization aims to predict 6 DoF camera poses in a pre-constructed 3D map from RGB images. 
Current localization pipelines can be divided into three categories:
1) End-to-end pose regression methods~\cite{kendall2015posenet,kendall2017geometric,brahmbhatt2018geometry,kendall2016modelling},
which directly regress absolute camera pose from a single RGB image. These methods are later shown to be closely related to image retrieval ~\cite{sattler2019understanding}.
2) Coordinate regression~\cite{schohn2000less, brachmann2017dsac, brachmann2019expert} methods regress the dense 3D scene coordinates of the query image and obtain the final camera pose by dense 2D-3D correspondences. Most of these methods are scene-specific and need to be trained for a new scene. 
3) Image matching based-methods. These methods first retrieve similar images from the database by image retrieval ~\cite{gordo2016deep, arandjelovic2016netvlad}, then establish sparse ~\cite{sarlin2019coarse} or dense ~\cite{rocco2018neighbourhood} correspondences between query images and retrieved database images to form the 2D-3D correspondences. Poses are finally solved by PnP algorithms. Current state-of-the-art methods for large-scale localization are mostly based on image matching. There are many works focus on local feature learning ~\cite{detone2018superpoint, dusmanu2019d2,luo2018geodesc,luo2019contextdesc,luo2020aslfeat}, sparse feature matching ~\cite{yi2018learning,zhang2019learning,sarlin2020superglue} and  dense matching ~\cite{rocco2018neighbourhood,rocco2020efficient} to improve the performance of image matching based localization methods.



However, image matching-based methods still suffer from significant performance drop if the database images have small overlaps with query images, which is common in large-scale environments. Our RenderNet eases this problem by generating additional virtual views.

\paragraph{View synthesis and pose estimation.} View synthesis aims to generate photo-realistic images of arbitrary novel view given 3D models, sparse views ~\cite{martin2020nerf} or point cloud ~\cite{sibbing2013sift, pittaluga2019revealing}. 
View synthesis has been shown useful in many pose estimation related-areas. 
Torii et al.~\cite{torii201524} show that image retrieval performance can be enhanced by enriching the database with novel views. 
Sibbing et al.~\cite{sibbing2013sift} render view from point clouds for visual localization and Zhang et al.~\cite{zhang2020reference} refine the estimated poses of localization by view synthesis. 

Our proposed RenderNet is inspired by the above applications of view synthesis. However, instead of direct rendering photo-realistic images,  RenderNet generates virtual views by directly generating the needed features used in the localization system, which will be shown can be more robust than the direct view synthesis approach.

%% file: methods.tex
\section{Visual relocalization by virtual viewpoints}

\begin{figure}
    \centering
    \includegraphics[width=0.5\textwidth]{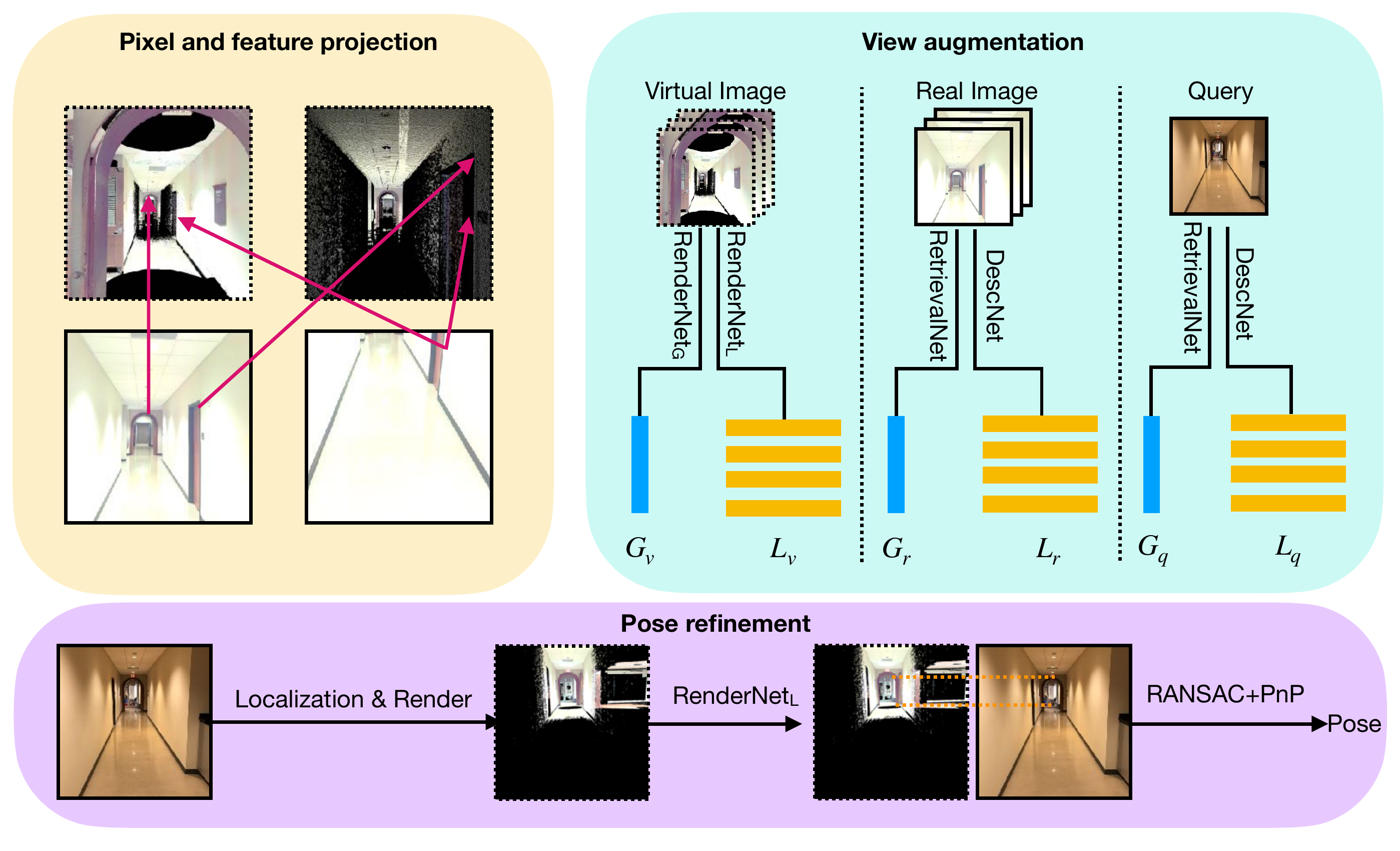}
    \caption{The overall pipeline of the proposed visual relocalization approach. 
    Top left: virtual views are generate by projected colors and local features from nearby real images. Top right: virutal views are used together with real database images for view augmentation. To be specific, RetrievalNet and DescNet are used for both query images and real database images, to generate global feature $G_q, G_r$ and local features $L_q, L_r$ for image retrieval and image matching respectively. $\text{RenderNet}_\text{G}$ and $\text{RenderNet}_\text{L}$ are used for the virtual view to generate corresponding global feature $G_v$ and local features $L_v$.
    Bottom: pose refinement to improve relocalization performance. The query pose is first estimated using RANSAC + PnP and further refined by taking the virtual view at the initial pose as the localization reference. 
    }
    \label{fig:pipeline}
\end{figure}

We first give an introduction to our relocalization method which uses virtual viewpoints for view augmentation and pose refinement. Then we present the details of RenderNet, which generates global and local features of virtual views used in the image retrieval and feature matching stage of the relocalization pipeline.

\subsection{Overview}
Our method follows the traditional paradigm which first retrieves similar images from database images by measuring the distance of their global features, and then estimates camera poses through 2D-3D correspondences from local feature matching. 
Besides, we generate virtual views to improve the image retrieval and image matching process.
With the generated novel view features, we design view augmentation and pose refinement stages to improve localization performance, as shown in Fig \ref{fig:pipeline}. 

\textbf{View augmentation stage.} 
In this stage, we enrich the database with rendered virtual views on a set of manually defined poses, which could have larger overlaps with queries than existing real database images. 
We first define a set of novel views around existing viewpoints and then generate their corresponding global and local features.
As shown in Fig. \ref{fig:pipeline}, for query images and real database images, we generate global feature vectors using off-the-shelf retrieval networks (RetrievalNet in Fig. \ref{fig:pipeline}), such as AP-GeM ~\cite{revaud2019learning}. In addition, interest points are detected and described using local feature networks (DesNet in Fig. \ref{fig:pipeline}), such as SuperPoint \cite{detone2018superpoint}. For virtual views, we generate their global feature vectors using our $\text{RenderNet}_\text{G}$ and their local features using $\text{RenderNet}_\text{L}$. 
In the retrieval stage, query images find the most similar top k database images by measuring the distance of global features. Real images and virtual images are treated equally here. 
After that, we can establish 2D-3D correspondences between query images and retrieved images through sparse feature matching. Finally, RANSAC + PnP is performed to estimate a coarse camera pose from established correspondences. 


\textbf{Pose refinement stage.} 
To refine the estimated camera pose, we render a new virtual view at the coarse pose estimated by the above localization pipeline, and then perform sparse matching and PnP pose estimation again. We generate only local features this time using $\text{RenderNet}_\text{L}$. As will be shown later, pose refinement can largely improve the pose accuracy, especially in the strict threshold region.

\begin{figure}
    \centering
    \includegraphics[width=0.5\textwidth]{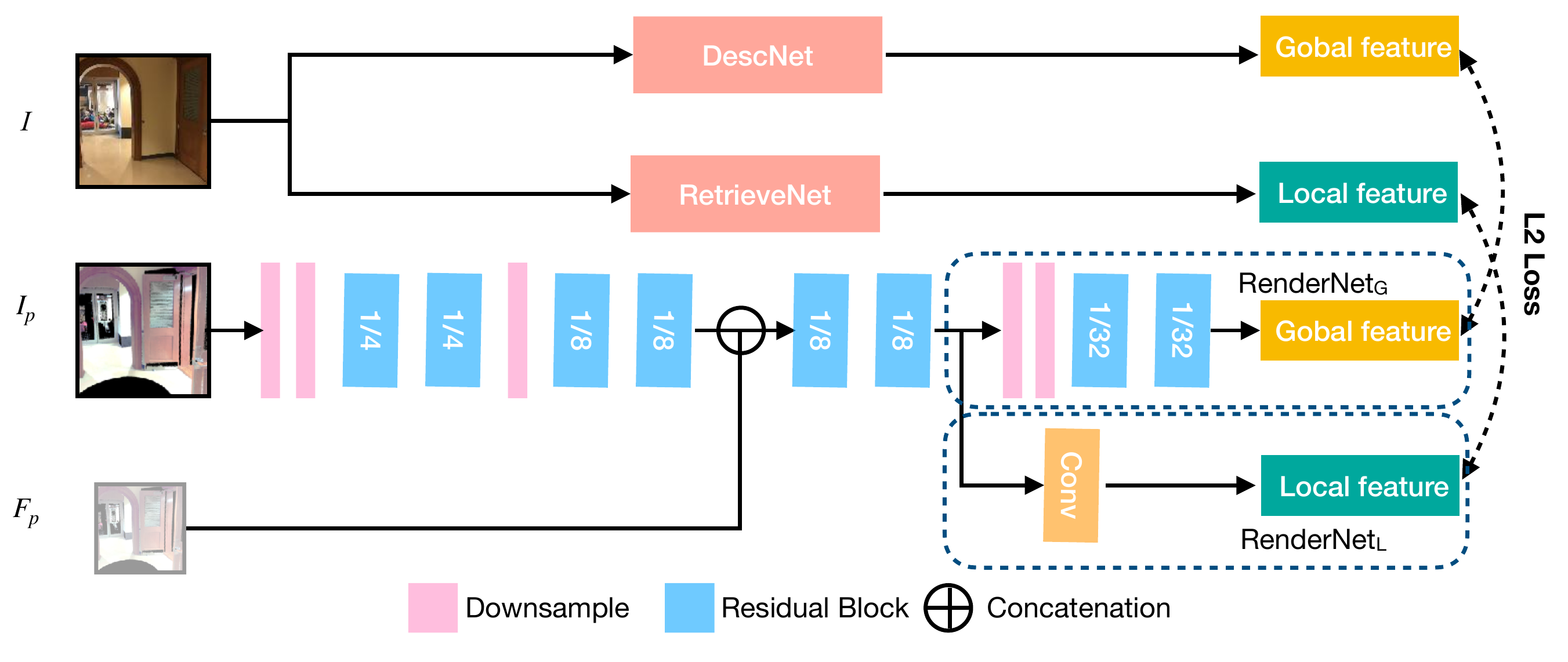}
    \caption{
    The training process of the RenderNet. RetrievalNet and DescNet are teacher models which take real RGB images as input, while $\text{RenderNet}_\text{G}$ and $\text{RenderNet}_\text{L}$, which take projected colors and local features as input, mimick the outputs of RetrievalNet and DescNet to generate global features and local features for virtual views.
    }
    \label{fig:training}
\end{figure}

\subsection{RenderNet}\label{sec:rendernet}

Here we present RenderNet to render an arbitrary novel virtual view. We directly generate the needed features rather than rendering a real image to avoid the artifacts when generating images.
More specifically, RenderNet contains two branch: 1) a  $\text{RenderNet}_\text{G}$ to generates a global feature for image retrieval. 2) a $\text{RenderNet}_\text{L}$ branch to generate a set of local features of keypoints in this view for sparse matching.
For the local branch, instead of re-detecting interest points in the virtual views, we reuse the projected keypoints from existing views and just generate revised descriptors that better describe this view.

\textbf{Input of RenderNet.} 
The input of RenderNet contains densely projected color and local feature information from nearby real database images, which is a $H \times W \times 3$ and $\frac{H}{8} \times \frac{W}{8} \times n$ tensor respectively, where $n$ is the channel dimension. Zeros are padded if no points are projected. 
Since local feature descriptors are good representations of image patches, we expect the encoded patch information can help resolve the artifacts in the projected color images.
We assume the database contains a coarse 3D model which can provide coarse depths to remove occluded points when projecting. A pre-trained DescNet is used as the local feature network, which is SuperPoint in our work.

\textbf{Output of RenderNet.} 
The $\text{RenderNet}_\text{G}$ branch aims to give a global representation of this virtual view. Here we adopt the widely used generalized mean pooling (GeM) layer to summarize this view. 
It outputs a $m$-d global feature vector which is used for image retrieval. In the retrieval stage, similarities are measured between real feature vectors and virtual feature vectors. 
So in order to ensure the distance comparison meaningful, we train the $\text{RenderNet}_\text{G}$ to minimize the distance between its output and the feature generated by RetrievalNet which takes a real image of this view, as follow,
\begin{equation}
argmin_{\phi}||\text{RenderNet}_\text{G}(I_{p}, F_{p}) - \text{RetrievalNet}(I) ||^2,
\end{equation}
where $\phi$ are the parameters of $\text{RenderNet}_\text{G}$, $I_{p}$ and $F_{p}$ are the projected colors images and projected features. $I$ is a real image of this view.

For the local features of this novel view, we use the projected keypoints and generate the revised local descriptions using $\text{RenderNet}_\text{L}$. 
Although the original local descriptors are learned to be invariant to viewpoint change, we find it is better to revise their representations for the current novel view, especially when viewpoints change largely.
It outputs a $n$-dimension feature map, then the local features of keypoints are interpolated from this feature map. Similarly, we train the $\text{RenderNet}_\text{L}$ to minimize the distance of its outputs and the local features generated by DescNet which takes a real image,
\begin{equation}
argmin_{\psi}||(\text{RenderNet}_\text{L}(I_{p}, F_{p}) - \text{DescNet}(I)) \cdot M ||^2,
\end{equation}
where $\psi$ are the parameters of $\text{RenderNet}_\text{L}$. A mask $M$ is used here to only consider the loss around the projected postions.

\textbf{Network architecture.} 
For real images, we use AP-GeM ~\cite{revaud2019learning} and SuperPoint ~\cite{detone2018superpoint} as the image retrieval and local feature extraction network respectively. 
As for our RenderNet, we use a modified ResNet18 as the backbone, which takes $H\times W \times3$ as input. As shown in Fig. \ref{fig:training}, RenderNet takes projected color as input and generates $\frac{1}{8}$ sized features. Then the projected $1/8$ sized local features are concatenated for further processing. Finally, the global feature and local feature maps are output by $\text{RenderNet}_\text{G}$ and $\text{RenderNet}_\text{L}$ branches respectively.
For the $\text{RenderNet}_\text{G}$ branch, after generating $1/32$ sized features, we use GeM pooling layer and a fully connected layer as the AP-GeM network ~\cite{revaud2019learning} to output a $m$d global feature. The $\text{RenderNet}_\text{L}$ branch generate a $1/8$ sized local feature maps as SuperPoint ~\cite{detone2018superpoint} using convolutional layesr.

\textbf{Training process.} As shown in Fig. \ref{fig:training}, we employ distillation training paradigms to learn the representation directly from teacher models.  Pre-trained RetrivalNet and DescNet serve as teacher models, which take input of real RGB images. For student model $\text{RenderNet}_\text{G}$ and $\text{RenderNet}_\text{L}$, their inputs are the projected colors and local features from other views. The output of teacher models serves as targets of $\text{RenderNet}_\text{G}$ and $\text{RenderNet}_\text{L}$ respectively. L2 loss is used in the training. 
We initialize the weights of head layers of $\text{RenderNet}_\text{G}$ and $\text{RenderNet}_\text{L}$ from their pre-trained teacher models,
and fix the weights during training.

%% file: experiment.tex
\section{Experiments}
\subsection{Experiment settings}
\label{sec:exp_setting}
\textbf{Training data.} We use Scannet~\cite{dai2017scannet} to train RenderNet. Scannet is an RGB-D video dataset consisting of 2.5M views in 1513 scenes with camera poses and dense depth maps provided. For each training image, we uniformly sample 4 nearby images and generate a virtual view by projecting color and local features to this view as illustrated in Sec. \ref{sec:rendernet}.  One training sample consists of a real image and a corresponding projected virtual image.

\textbf{Testing data.} We evaluate our method in the public available dataset InLoc~\cite{taira2018inloc}. InLoc contains 2 large-scale indoor scenes, covering around 10000 $m^2$ areas, which are captured using a high-end 3D scanner. 
It provides 4680 RGB database images extracted from original RGBD panoramic images and 356 RGB query images with ground truth pose. We use standard evaluation metrics accuracy under threshold of ((0.25m, 2°) / (0.5m, 5°) / (5m, 10°)). 
Before testing, our RenderNet models are finetuned the Inloc dataset using their database images.

\textbf{Implementation details.} 
We train the RenderNet for 30 epochs with a batch size of 24 on four RTX 2080TI GPUs, 80000 pairs are randomly sampled in each epoch. 
When testing, global features are whitened using Principal Component Analysis (PCA) trained from descriptors of all real and virtual images. 
Local features are matched using SuperGlue~\cite{sarlin2020superglue}, a transformer-based sparse feature matcher, which is trained with SuperPoint local features. Top 40 database images are matched for the Inloc dataset.

\begin{table}[]
\small
\centering
\scalebox{0.8}{

\begin{tabular}{c|ccc|ccc}
\Xhline{2\arrayrulewidth}

                                                          & \multicolumn{3}{c|}{DUC1}                                 & \multicolumn{3}{c}{DUC2}                                 \\ \Xhline{2\arrayrulewidth}

InLoc\cite{taira2018inloc}         & 40.9                      & 58.1          & 70.2          & 35.9                      & 54.2          & 69.5          \\ \hline
DensePNV\cite{taira2019right}      & 40.9 & 63.6          & 71.7          & 42.7 & 61.8          & 71.0          \\ \hline
R2D2\cite{humenberger2020robust}   & 41.4                      & 60.1          & 73.7          & 47.3                      & 67.2          & 73.3          \\ \hline
D2Net\cite{dusmanu2019d2}          & 42.9                      & 63.1          & 75.3          & 40.5                      & 61.8          & 77.9          \\ \hline
NCNet\cite{rocco2018neighbourhood} & 43.4                      & 64.6          & 77.8          & 45.0                      & 62.6          & 73.3          \\ \hline
HLoc\cite{sarlin2019coarse}        & 46.5                      & 65.7          & 77.8          & 51.9                      & 72.5          & 79.4          \\ \hline
SparseNCNet\cite{rocco2020efficient}                                               & 47.0                      & 67.2          & 79.8          & 43.5                      & 64.9          & 80.2          \\ \hline
SuperGlue + Patch2Pixel\cite{zhou2020patch2pix}                                   & 50.0                      & 68.2          & 81.8          & 57.3                      & 77.9          & 80.2          \\ \hline
RLOCS\cite{fan2020visual}                                                     & 47.0                      & 71.2          & 84.8          & 58.8                      & 77.9          & 80.9          \\ \hline
LoFTR\cite{sun2021loftr}                                                     & 47.5                      & 72.2          & 84.8          & 54.2                      & 74.8          & \textbf{85.5} \\ \hline
Ours                                                      & \textbf{58.6}             & \textbf{77.8} & \textbf{89.4} & \textbf{74.8}             & \textbf{82.4} & \textbf{85.5}          \\ \Xhline{2\arrayrulewidth}

\end{tabular}
}
\caption{Performance comparison in terms of accuracy under threshold of (0.25m, 2°) / (0.5m, 5°) / (5m, 10°) in InLoc benchmark. 
}
\label{tab:inloc}
\end{table}

\subsection{Results of InLoc}
\label{sec:exp_inloc}
\begin{figure*}
\includegraphics[width=\textwidth]{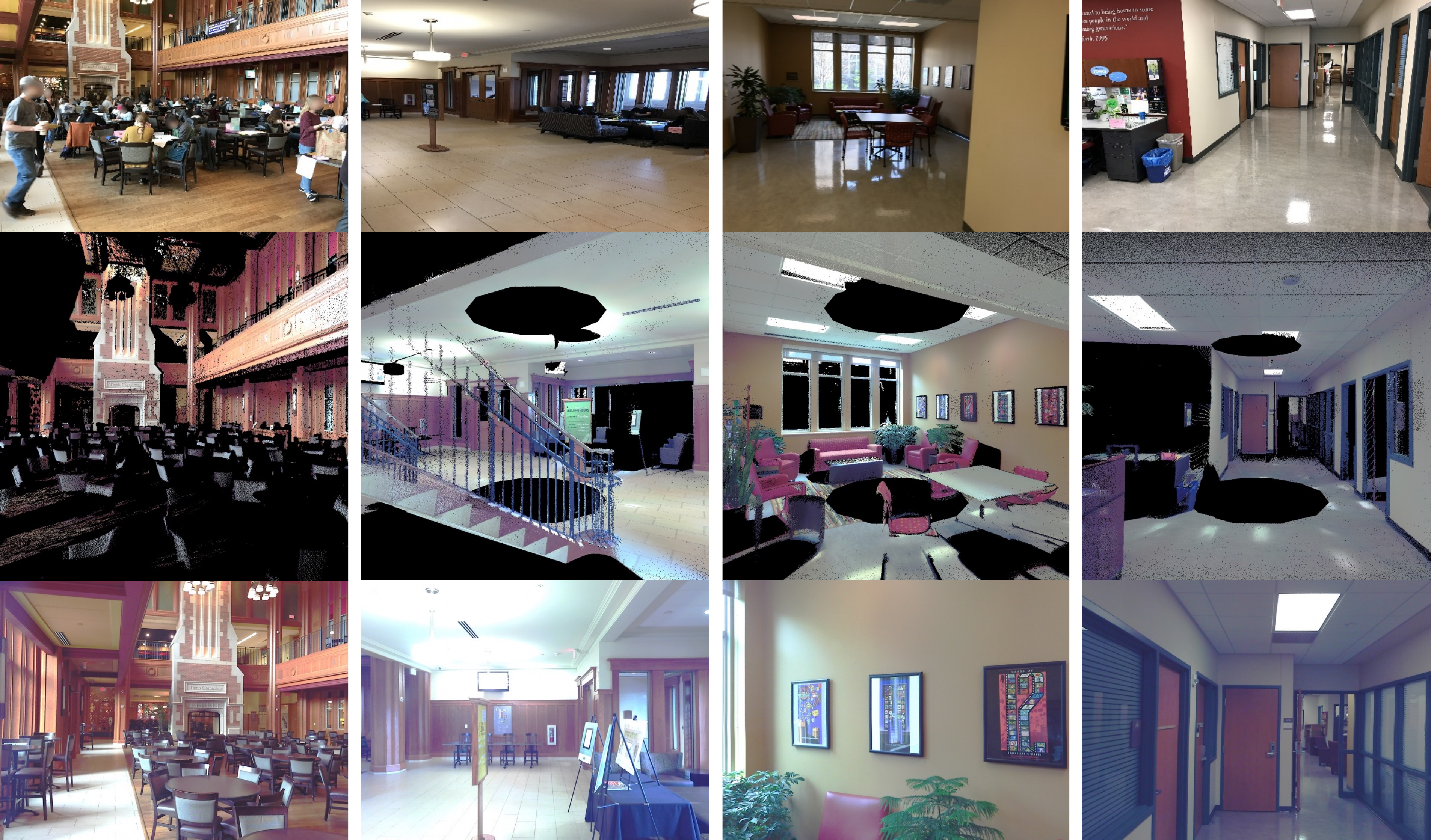}

\caption{
Rendered virtual images can have larger overlaps with queries than existing real database images in the Inloc dataset. Top: query images. Middle: most similar virtual images. Bottom: most similar real database image.
}\label{vis}
\end{figure*}

For the inloc dataset, we generate novel virtual views by moving the camera center horizontally to four positions in the front, back, left, and right respectively.
At each position, 12 images are rendered by rotating the camera every 30° horizontally. The moving distance is 5m and 2.5m for DUC1 and DUC2 respectively, finally resulting in 5556 valid virtual images. 
In Fig. \ref{vis}, we show the virtual views retrieved by $\text{RenderNet}_\text{G}$ for different query images. The virtual views shown in the figure are generated by projection from nearby views.
It can be identified that the added views have larger overlaps with query images than the real database image, which makes it easier to be retrieved and matched. 
For quantitative evaluation, we compare RenderNet with various baselines in terms of accuracy under threshold of (0.25m, 2°) / (0.5m, 5°) / (5m, 10°), such as sparse image matching based methods ~\cite{dusmanu2019d2,humenberger2020robust,fan2020visual,zhou2020patch2pix} and dense image matching based methods ~\cite{rocco2018neighbourhood,rocco2020efficient,sun2021loftr}.
The results are shown in Tab. \ref{tab:inloc}, our method gives much better results than the other methods, especially under the threshold of 0.25m. Compared with very recent state-of-the-art LoFTR ~\cite{sun2021loftr} which is a transformer-based dense image matching method, we achieve improvements on most metrics, which are 11.1\%/5.6\%/4.6\% and 20.6\%/7.6\%/0.0\% on DUC1 and DUC2 respectively. 

\subsection{Ablation Study}
\begin{table}[]
\small
\centering
\scalebox{0.85}{
\begin{tabular}{c|ccc|ccc}
                            & \multicolumn{3}{c|}{DUC1}                     & \multicolumn{3}{c}{DUC2}                      \\ \hline
Baseline                    & 50.0          & 69.7          & 82.3          & 48.9          & 68.7          & 73.3          \\
VA (RenderImageNet)           & 45.5          & 70.2          & 84.3          & 55.0          & 74.8          & 79.4          \\
VA (RenderNet)              & 54.0          & 75.3          & 88.9          & 64.9          & \textbf{84.0}          & \textbf{85.5} \\ \hline
VA + PR (Full RenderNet)               & \textbf{58.6} & \textbf{77.8} & \textbf{89.4} & \textbf{74.8} & 82.4 & \textbf{85.5} \\ \hline
VA (RenderNet w/o local)    & 42.4          & 69.2          & 80.8          & 64.1          & 80.9          & 83.2    \\
VA (RenderNet w/o finetune) & 47.5          & 71.2          & 85.4          & 56.5          & 75.6          & 78.6  
     
\end{tabular}
}
\caption{Ablation about view augmentation (VA) and pose refinement (PR) in the Inloc dataset. 
Results are given in terms of accuracy under threshold of (0.25m, 2°) / (0.5m, 5°) / (5m, 10°). 
}\label{tab:ablations}
\end{table}

\label{sec:ablation}

\textbf{View augmentation}. We first show the effectiveness of view augmentation. Our baseline is a modified HLoc ~\cite{sarlin2019coarse}, which uses DIR ~\cite{revaud2019learning} network for image retrieval. This baseline achieves higher results than the original HLoc in DUC1.
In Table. \ref{tab:ablations}, the notation VA means that we use view augmentation in the localization pipeline.

As is shown, view augmentation via RenderNet in Tab. \ref{tab:ablations}) can largely boost the performance, achieving an improvement of  4.0\%/5.6\%/6.5\% and 16.0\%/15.3\%/12.2\% in DUC1 and DUC2 respectively. In order to compare with the view synthesis method which generates a real image for localization, we design a RenderImageNet which also takes projected image and local features as input. The RenderImageNet is learned to synthesize a real image of the input view similar to our RenderNet. It is also trained on Scannet and finetuned on Inloc dataset. As shown in the Table. \ref{tab:ablations}, our RenderNet can achieves much better than VA (RenderImageNet in Tab. \ref{tab:ablations}), 8.5\%/5.1\%/4.6\% and 9.9\%/9.2\%6.1\% higher on DUC1 nad DUC2 respectively. This demonstrates that directly rendering the needed features can be more robust than synthesizing images. Compared with DUC1, the improvement of RenderNet in DUC2 is higher. This is because DUC2 is more narrow and scene appearance changes larger when viewpoint changes, thus rendering additional views is more helpful for DUC2. 

\textbf{Effectiveness of $\text{RenderNet}_\text{L}$}. In order to verify the effectiveness of $\text{RenderNet}_\text{L}$, we directly use the projected local feature descriptors without revising the features for the current view. Although SuperPoint are learned to be invariant to view changes, the revised features learned by our $\text{RenderNet}_\text{L}$ can achieve higher results in large viewpoint change case, with an improvement of 8.1\% and 2.3\% under the 5m, 10° threshold, comparing to the VA + RenderNet w/o local shown in Table. \ref{tab:ablations}.

\textbf{Effectiveness of finetune}. In Table . \ref{tab:ablations}, we also report the results without finetuning RenderNet on the target dataset Inloc. It can also improve the localization performance. After finetune, RenderNet can further achieve 3.5\% and 6.9\% higher results under the 5m, 10° threshold. Since the finetune can be applied offline and does not increase the cost of online relocalization system, we recommend a finetune on the target dataset for better performance.

\textbf{Pose refinement}. We then validate the effectiveness of pose refinement. In Table. \ref{tab:ablations}, Full RenderNet means we adopt both view augmentation and pose refinement. Using both rendered virtual images and real images can largely improve the pose accuracy.
As is shown, when using pose refinement, the accuracy under 0.25m, 2° can be further increased by 4.6\% and 9.9\% compared to only using view augmentation. 
We can see that view augmentation improves all the metrics while pose refinement mainly increases the accuracy of threshold 0.25m by a large margin. The reason is that view augmentation makes it more likely for query images to retrieve views with large overlaps. On the other hand, pose refinement fine-tunes the coarse poses to a more accurate one. Combining these two approaches, the RenderNet can boost the performance by 8.6\%/8.1\%/7.1\% and 25.9\%/13.7\%/12.2\%.

%% file: conclusion.tex
\section{Conclusion}
In the work, we propose RenderNet for relocalization in large scale indoor environments, which generates virtual views for view augmentation and pose refinement without requiring high-quality 3D models. It consists of $\text{RenderNet}_\text{G}$ to generate global features and $\text{RenderNet}_\text{L}$ to generate local descriptors for virtual novel views, which are then used in the relocalization pipeline for image retrieval and image matching. These added novel views help retrieve the right place and establish more feature correspondences.
We show in experiments that our method can largely improve the relocalization performance in the large-scale indoor dataset Inloc.